\documentclass[sigconf,nonacm]{acmart}
\AtBeginDocument{%
  }

\usepackage[most]{tcolorbox}

\usepackage{xcolor}
\usepackage{tikz}
\usetikzlibrary{matrix}
\usepackage{graphicx}

\definecolor{warmivory}{HTML}{F3ECE1}
\definecolor{deepnavy}{HTML}{192735}
\definecolor{palesage}{HTML}{C9CCC2}
\definecolor{mutedgold}{HTML}{E0C9A9}
\definecolor{softbeige}{HTML}{E7DDCD}
\definecolor{olivesage}{HTML}{767C6D}

\newtcolorbox{layer}[1][]{
  width=.96\columnwidth,
  boxrule=.5pt,
  arc=2pt,
  left=4pt, right=4pt, top=4pt, bottom=4pt,
  before skip=2pt, after skip=2pt,
  #1
}

\begin{document}

\title{Beyond Tasks: A Vision for Reproducing an Animal-like Behavioral Substrate Using Modern Robot Learning Techniques}

\author{Samiyuru Menik}
\affiliation{%
  \department{School of Computing}
  \institution{University of Georgia}
  \city{Athens}
  \state{GA}
  \country{USA}
}
\email{sami.menik@uga.edu}

\author{Hemadri Jayalath}
\affiliation{%
  \department{School of Computing}
  \institution{University of Georgia}
  \city{Athens}
  \state{GA}
  \country{USA}
}
\email{hemadri.jayalath@uga.edu}

\begin{abstract}

Recent advances in robot learning have produced increasingly capable embodied agents. Yet comparatively less attention has been given to a more basic form of competence that animals exhibit continuously: the ability to remain situated, responsive, and behaviorally coherent as physical, environmental, and social demands change over time. We propose the \emph{ethological behavioral substrate} as a conceptual lens for studying this form of competence in artificial agents. Rather than treating these behaviors that animals exhibit as a set of isolated skills, we argue that their continual coordination under competing demands constitutes an important and underexplored target for modern robot learning. We further propose robotic animal companions as a useful research setting for studying sustained interaction and adaptation in human-centered environments. Such systems provide an opportunity to investigate how social behavior, memory, and continual learning develop over long periods of interaction. This perspective motivates further investigation of how such persistent behavioral competence may complement higher-level capabilities in embodied agents.

\end{abstract}




\begin{teaserfigure}
    \centering
    \includegraphics[width=0.49\textwidth]{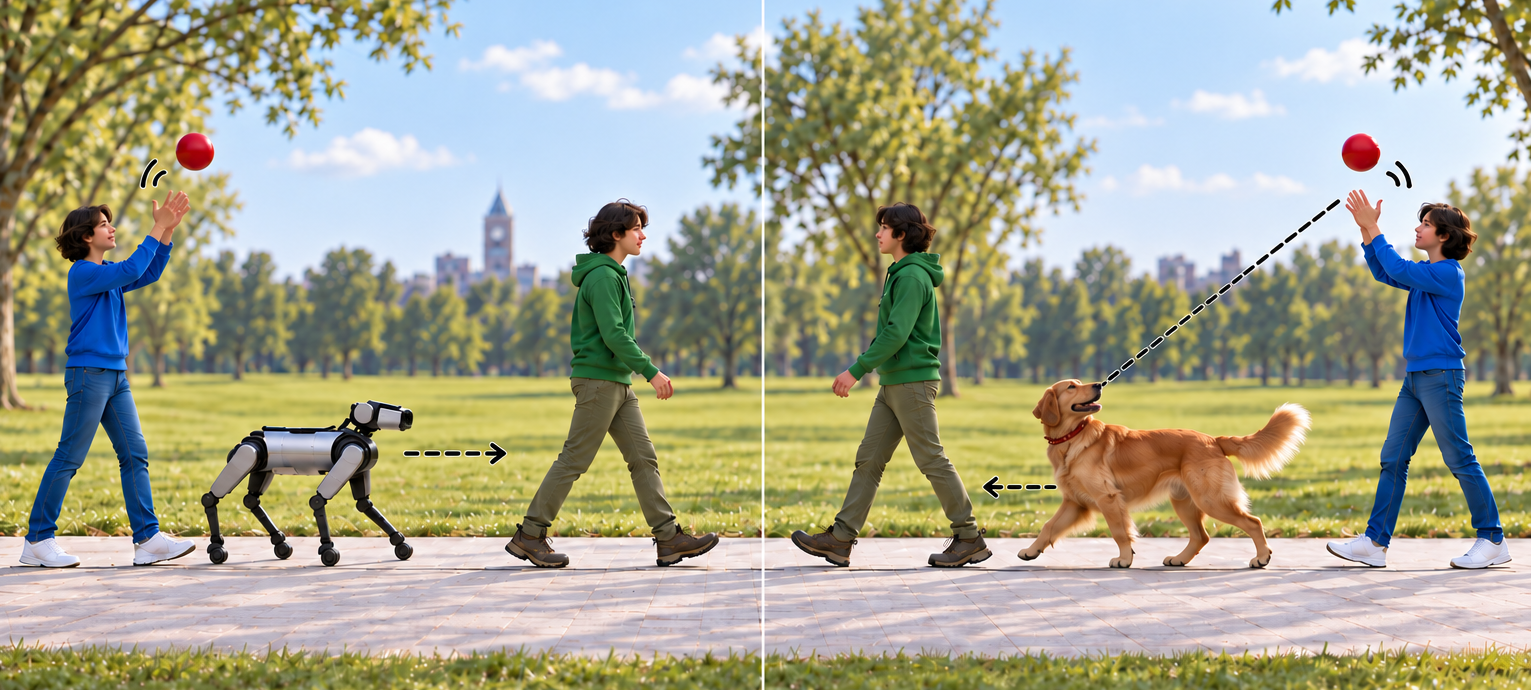}%
    \hfill
    \includegraphics[width=0.49\textwidth]{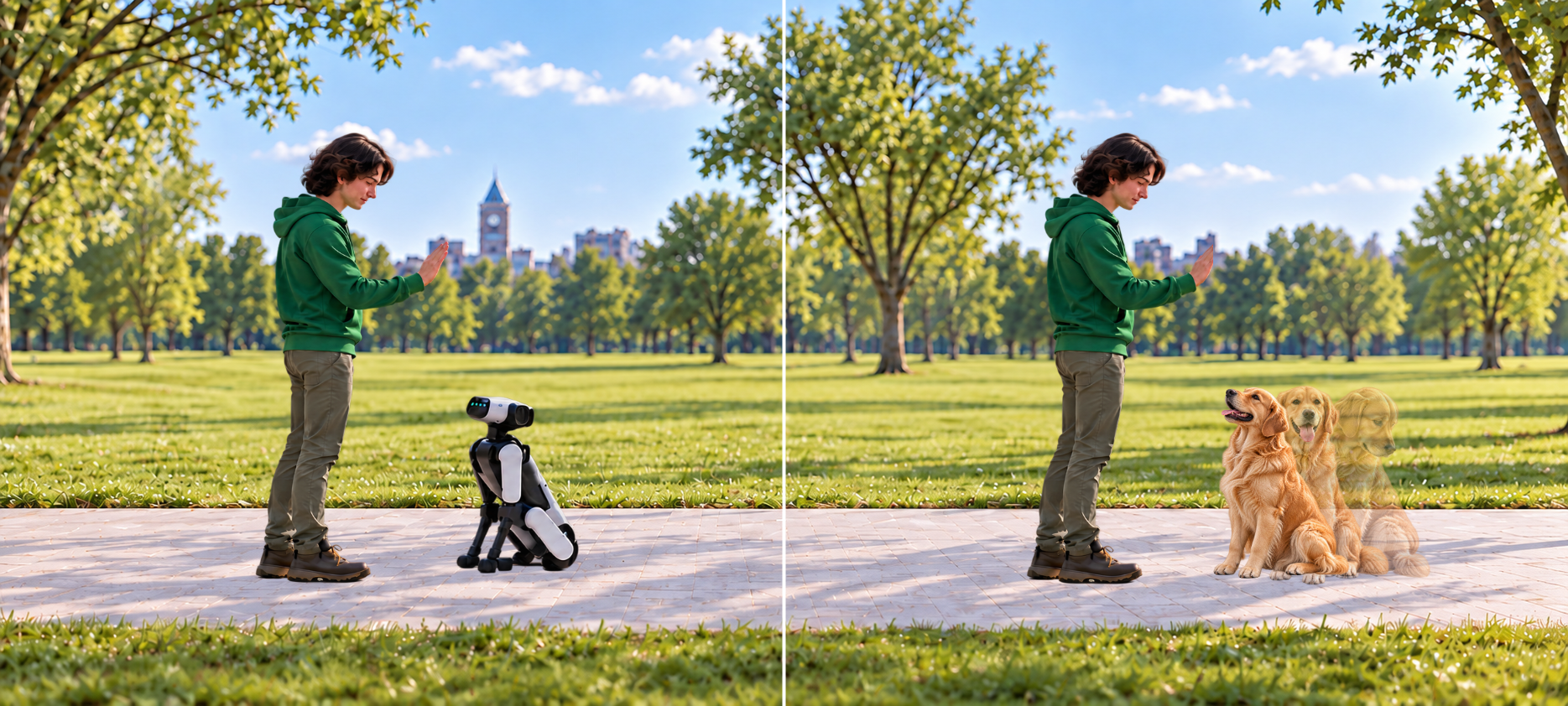}
    \caption{Illustration of how animal behavior remains responsive to the broader physical and social context even while pursuing an ongoing activity. In the left scene, the agent follows an externally directed activity while continuing to attend and respond to surrounding environmental stimuli. In the right scene, behavior is not determined solely by the immediate instruction: the agent may orient, pause, engage, or disengage and respond to other relevant physical and social cues. We explore the idea of an ethological behavioral substrate that continually coordinates behavior as conditions and competing demands change over time.}
    \label{fig:teaser}
\end{teaserfigure}

\maketitle

\pagestyle{plain}

\begingroup
\renewcommand{\thefootnote}{}
\footnotetext{Copyright \textcopyright\ 2026 The Authors.
This work is licensed under a Creative Commons Attribution 4.0
International License (CC BY 4.0):
\url{https://creativecommons.org/licenses/by/4.0/}.}
\addtocounter{footnote}{-1}
\endgroup

\section{Introduction}

Physical intelligence is emerging as a natural next step for agentic AI seeking to extend the capabilities of digital AI into the physical world. Recent advances in robot learning have shown increasing potential for general-purpose behavior. Rather than being specialized for a single task, modern robot learning systems can acquire skills that generalize across tasks and environments. These capabilities include language-conditioned behavior and a growing range of embodied skills ~\cite{zitkovich2023rt2,ghosh2024octo}. Alongside this progress an interesting question is underexplored: can modern robot-learning systems acquire and continually adapt a coordinated repertoire of animal-like behaviors during sustained physical and social interaction?

Many animals exhibit a persistent repertoire of primal behaviors that enables continuous and adaptive interaction with their surroundings. This repertoire encompasses attentional orienting, exploration, hazard avoidance, boundary maintenance, and the prioritization of competing stimuli, together with interaction with other agents. Domestic animals further display sustained patterns of social behavior toward humans, such as approaching, following, maintaining proximity, responding to human attention, and disengaging when appropriate.

Taken together, these behaviors form a basic functional repertoire for continuous embodied interaction, which we refer to as the ethological behavioral substrate. Its defining feature is not the presence of individual capabilities in isolation, but their coordinated expression over time as environmental and social conditions change. Such coordination enables an agent to adapt locally, maintain its safety and physical integrity, and sustain forms of coexistence and symbiotic interaction with humans and other animals.

This perspective raises a central question for robot learning: can modern learning systems acquire and coordinate a persistent repertoire of animal-like behaviors that supports adaptive, context-sensitive interaction over time? A central challenge is not simply to acquire individual behaviors, but to recognize and resolve competing behavioral demands in ways that remain compatible with the surrounding physical and social environment. We examine the prospects for developing such an ethological behavioral substrate in artificial agents and identify the main research challenges involved. We also consider the broader scientific and practical impact of this research direction for building embodied agents capable of sustained and adaptive coexistence in the physical world.

\section{Vision}

We posit that basic animal-like behavioral competence is a promising research target for modern robot learning and a useful complement to current efforts toward increasingly general robotic intelligence.
Our central hypothesis is that capabilities such as attention, exploration, stimulus prioritization, spatial regulation, mobility, and elementary social interaction can constitute a reusable behavioral substrate for embodied agents. These capabilities are valuable in their own right, but may also provide a foundation for acquiring and expressing more sophisticated behavior.

\begin{figure}[t]
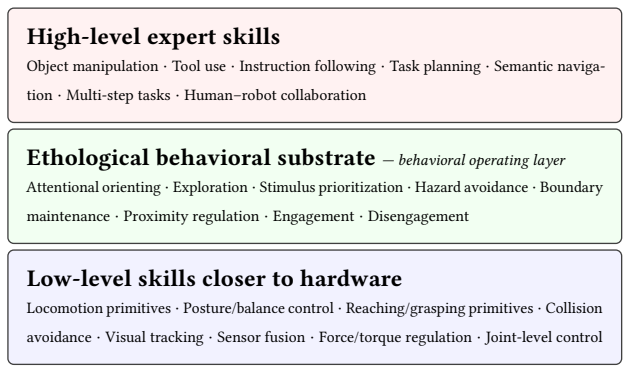

\centering

\begin{layer}[colback=red!5]
\textbf{High-level expert skills}\\[-1pt]
{\scriptsize
Object manipulation $\cdot$ Tool use $\cdot$ Instruction following $\cdot$
Task planning $\cdot$ Semantic navigation $\cdot$ Multi-step tasks $\cdot$
Human--robot collaboration}
\end{layer}

\begin{layer}[colback=green!5]
\textbf{Ethological behavioral substrate}
{\scriptsize\itshape --- behavioral operating layer}\\[-1pt]
{\scriptsize
Attentional orienting $\cdot$ Exploration $\cdot$ Stimulus prioritization
$\cdot$ Hazard avoidance $\cdot$ Boundary maintenance $\cdot$ Proximity regulation $\cdot$ Engagement $\cdot$ Disengagement}
\end{layer}

\begin{layer}[colback=blue!5]
\textbf{Low-level skills closer to hardware}\\[-1pt]
{\scriptsize
Locomotion primitives $\cdot$ Posture/balance control $\cdot$
Reaching/grasping primitives $\cdot$ Collision avoidance $\cdot$
Visual tracking $\cdot$ Sensor fusion $\cdot$ Force/torque regulation
$\cdot$ Joint-level control}
\end{layer}

\caption{A layered view of competence of robotic agents that integrate into physical and social environments.}
\label{fig:ethological-layers}
\end{figure}

We conceptualize this substrate as a mid-level behavioral operating layer: a set of mechanisms that continuously regulate how an embodied agent attends, moves, explores, prioritizes stimuli, and responds to its surroundings. This notion is a functional abstraction rather than a prescribed architecture. It could be realized within a unified vision-language-action (VLA) model, a hierarchical policy, or another learning architecture.

The broader research objective is to reproduce this foundational competence in increasingly capable robots and investigate whether preserving it alongside higher-level capabilities improves persistent interaction, adaptation, and learning in the physical and social world.

An evolutionary perspective provides additional motivation. Basic behavioral and regulatory capacities predate specialized nervous-system organization, which appears to have evolved in organisms already capable of sustained interaction with their environments~\cite{Budd2015}. This observation motivates a question for robotics: does competence in basic persistent behaviors provide a useful foundation for more sophisticated embodied behavior?

\section{Research Directions}

One of the important research directions is to investigate minimum viable hardware platforms for ethological robot agents and determine which sensorimotor capabilities are sufficient to support animal-like interactions. Candidate capabilities could include biologically inspired stereo vision, binaural audition, inertial sensing for balance and orientation, omnidirectional locomotion, and speakers for active communication, with tactile sensing as an additional modality.

Promising vision-language-action (VLA) models learn robot actions from large collections of demonstrations, alongside pretrained visual and language representations~\cite{kim2025openvla}. Following that path, the initial data collection could focus on naturally occurring sensorimotor coordination demonstrations. One particularly interesting class of behaviors involves one sensory modality guiding information gathering through another, such as orienting toward an auditory event and subsequently using vision to acquire additional information and select a response. It would also be useful to study behavioral regularities such as where animals position themselves while idle, which paths they select through an environment, what stimuli elicit exploratory or curiosity-related behavior, and how human–animal interaction patterns unfold over time. Instrumented harnesses and other minimally intrusive sensing systems are worth exploring as mechanisms for collecting synchronized multimodal behavioral data from animals in naturalistic settings~\cite{ehsani2018dogs}.

Another research question concerns how to learn across the many dimensions of perception, action, affect, context, memory, and social interaction. Learning across these dimensions jointly could be infeasible. Instead of learning a joint distribution using a single model, modular approaches~\cite{menik2023towards, menik2021robust} should be investigated in which related behavioral or sensorimotor components are learned separately and subsequently integrated through conditional models or other coordination mechanisms. For example, specialized models for locomotion, orienting, social engagement, exploratory behavior, and affect-related responses could be studied individually before examining how their conditional dependencies can produce coherent behavior at the system level.

The computational architecture introduces a related set of systems questions. It is useful to investigate architectures in which high-frequency processes, such as balance, locomotion, and immediate sensorimotor responses, execute locally on the robot, while lower-frequency processes, such as planning, long-term memory, social reasoning, and computationally intensive learning, may execute on separate computational resources~\cite{ichnowski2023fogros2}. Optimal ways to divide computation across these timescales, maintain behavioral coherence, and manage communication latency remain an important area for study.

Robot agents with an ethological behavioral substrate should be studied as systems that integrate into human-centered environments and support sustained interaction resembling some aspects of human relationships with companion animals. Relevant properties to investigate include a recognizable identity, continuity of shared experiences, adaptation to an individual owner, autonomous initiative, stable yet evolving behavior, attachment-like responses, shared routines, opportunities for caregiving, and continuity across periods of absence~\cite{friedman2003hardware}. These properties should not be treated as a fixed catalog of manually programmed behaviors. Instead, it is worth studying whether and how they can emerge through embodied learning, continual adaptation, long-term episodic and semantic memory, and interaction histories conditioned on the human and household context. End-to-end learning is one possible approach, but its ability to produce these properties should be evaluated rather than assumed.

This may also provide a controllable and reproducible platform for studying sustained human-robot interaction. Their behavior, embodiment, and interaction policies can be systematically manipulated, enabling researchers to examine how users interpret and respond to adaptive behavior, form expectations, develop trust or attachment, and personalize their interactions with a robot over time.

Benchmarks are important for defining the expected behavior and for comparing different approaches to developing them. Evaluation should address not only whether individual behaviors can be produced, but also whether they can be coordinated appropriately as physical and social conditions change over time. This includes assessing behavior selection under competing demands and adaptation to environmental changes as well as interaction with humans and other agents. Long-term evaluation should also examine whether behavior remains coherent across repeated interactions and changing social contexts. Developing benchmark environments and metrics for these properties is therefore an important research direction.

These questions motivate exploration of online continual and lifelong learning, imitation learning and behavioral cloning, memory-augmented architectures, autoregressive sequence modeling, reinforcement learning, and affective computing within long-term embodied agents. Of particular interest is the problem of learning continuously from humans during extended deployment rather than treating training and deployment as completely separate phases~\cite{liu2023sirius}. Longitudinal deployment could also provide opportunities to study new forms of behavioral data collection, while raising important research questions concerning privacy-preserving sensing, storage, memory, and adaptation.

\section{Robotic Agents in Human-Centered Environments}

\begin{figure}[ht]
    \centering
    \includegraphics[width=0.47\textwidth]{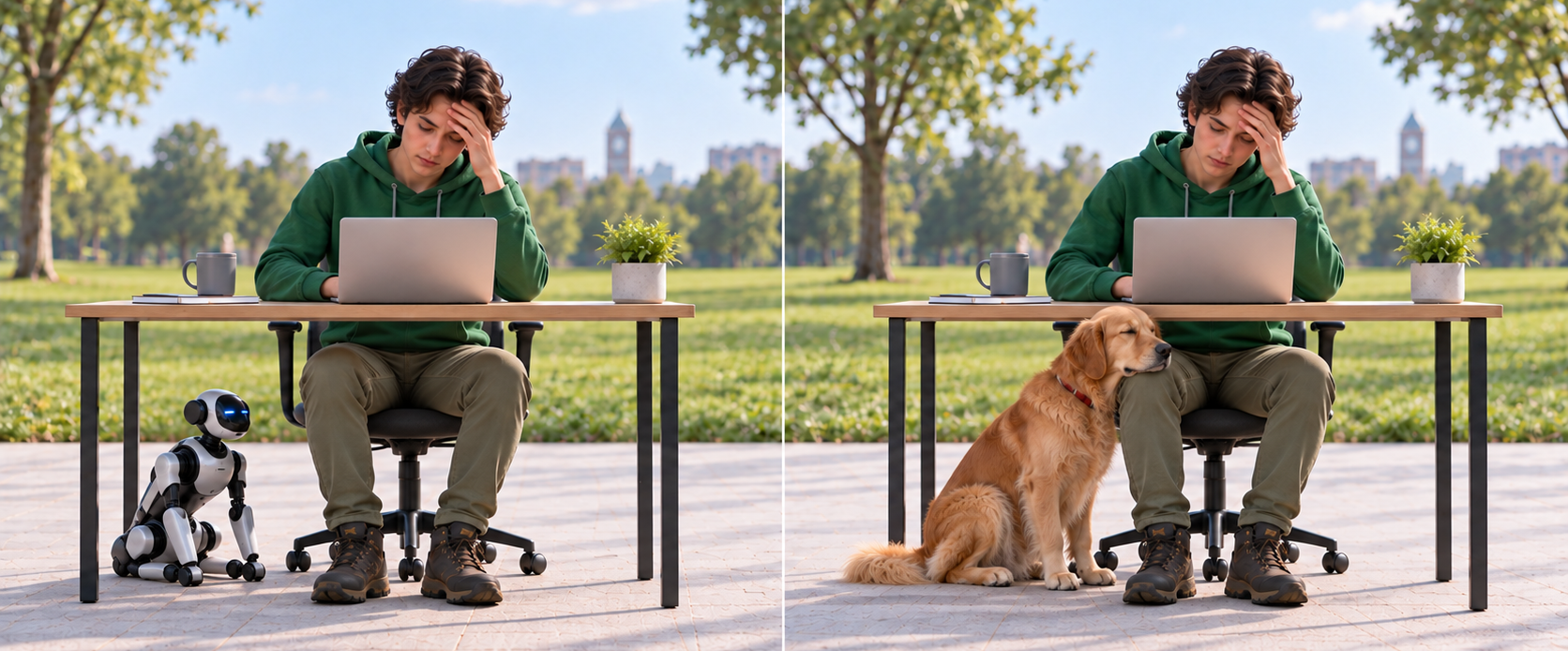}
    \caption{Affective interaction in human-centered environments. Similar to companion animals, robotic agents may perceive and respond to human affect through multimodal behavioral cues, while expressing their own internal states through movement, posture, sound, and other behaviors to support socially meaningful and adaptive interaction.}
    \label{fig:robot}
\end{figure}

One promising application of this research direction is the development of robotic agents that can sustain adaptive interaction within human-centered environments. A pet-companion context aligns naturally with the behavioral substrate we propose for several reasons. In particular, a pet-like agent can accommodate substantial behavioral variability without necessarily violating user expectations. Different strategies, habits, preferences, and personality-like behavioral patterns can remain plausible within the interaction. This may make adaptive and non-deterministic behavior easier to accommodate than in systems expected to behave with human-like consistency or task-oriented precision. Pet-like embodiments may also place fewer expectations on high-skill or precision tasks than humanoid robots, potentially reducing perceptual and behavioral mismatches. As an example, reliable precision manipulation remains challenging for contemporary robot-learning systems, but such capabilities are not central to the form of interaction considered in this context. More importantly, because interaction with a companion robot need not be primarily productivity-oriented, it provides a comparatively low-task-pressure setting for studying spontaneous interaction, social signaling, play, attachment, and the formation of long-term behavioral expectations.
When considering the societal impact, we see particular value in a robot pet companion that can sustain a persistent repertoire of coordinated, context-sensitive behaviors, rather than relying primarily on isolated or predefined interactions. 

Animal companionship can provide social interaction, routine, affection, and opportunities for engagement, and has been associated with benefits related to loneliness and social isolation \cite{kretzler2022pet}. Conventional pet ownership, however, entails substantial commitments involving care, exercise, hygiene, medical attention, time, and financial resources, and may not be feasible for individuals with allergies or other constraints. Robotic animal companions could provide some forms of companionship and interaction without requiring the full set of responsibilities associated with caring for a living animal.
Such systems may also be useful in contexts involving populations with particular interaction or accessibility needs. For older adults and people with cognitive or mobility limitations, a robotic companion could combine persistent social interaction with assistive functions such as reminders, monitoring, or caregiver notifications. For children, robotic companions have been explored as tools for encouraging physical and social interaction, supporting communication, and providing structured opportunities to learn caregiving and appropriate interaction with animals \cite{melson2009children}. In healthcare and rehabilitation settings, they have also been investigated as a means of supporting engagement, distraction, and emotional comfort during treatment \cite{moerman2019social}. These settings provide concrete environments in which the proposed behavioral substrate could be evaluated through sustained interaction over time.

\subsection{Affective interaction}

Affective responsiveness is an important component of companionship. Humans infer affective states from facial expressions, body language, vocal characteristics, and contextual cues, while companion animals such as dogs can also discriminate aspects of human affect and modify their behavior accordingly \cite{lange2022reading,albuquerque2023dogs}. These observations motivate the study of affect as part of a robotic companion's broader behavioral repertoire.

Affective computing methods can enable a robot to infer aspects of a user's affective state from multimodal signals such as facial expressions, body movement, speech, and physiological measurements \cite{spezialetti2020emotion, hegde2025emotions}. The robot can, in turn, convey aspects of its internal state through movement, posture, sound, or other observable cues, potentially making its responses more interpretable and socially legible \cite{park2022empathy}.
Of particular interest to this research direction is the integration of affective expression into learned behavior rather than treating it as a separate set of scripted animations. For example, EmoLo learns emotion-inspired locomotion styles within a reinforcement-learning policy \cite{kobayashi2026emolo}. Olaf uses animation-guided reinforcement learning to learn physically feasible whole-body motions that preserve an animator-designed expressive character style, rather than explicitly learning emotional or affective states from human labels \cite{muller2026olaf}. Extending these ideas would allow affect perception and expression to influence the same behavioral substrate that governs the robot's other actions, enabling researchers to study how affective signals interact with adaptation, action selection, and behavioral continuity.

\subsection{Data flywheel}

A persistent repertoire of coordinated, context-sensitive behaviors may provide a particularly rich setting for continual and interactive learning. Unlike a task-oriented robot that may remain idle between assignments, an agent that continuously attends to, explores, and engages with its surroundings can generate sensorimotor and interaction data over extended periods. These experiences can in turn support further learning and adaptation. More generally, prior work has shown how feedback generated during system operation can be analyzed to continually improve deployed machine-learning capabilities \cite{jayalath2022enhancing,jayalath2023continual}. This creates a potential feedback loop between persistent behavior, experience collection, and capability improvement.

A robotic animal companion may be particularly well suited to this process because users can naturally provide feedback through praise, correction, demonstration, preference expression, verbal instruction, and repeated interaction. Its continued presence in the same environment also allows the system to accumulate experience across different situations and behaviors. Because a robotic animal  can engage in activities such as locomotion, play, following, fetching, monitoring, object interaction, and social interaction, experience acquired through one behavior may also provide training signals or representations useful for others. In this way, persistent engagement can turn otherwise idle periods into opportunities for data collection and continual learning.

\subsection{Safety and privacy}
Persistent robotic companions also raise safety and privacy questions that differ from those of robots used only for discrete tasks. A robot operating continuously in a home may observe sensitive information through cameras, microphones, localization systems, and other sensors \cite{levinson2024our}. The same perceptual capabilities that support richer interaction and adaptation can therefore increase the amount and sensitivity of information available to the system \cite{levinson2024our,eick2020enhancing}.
This creates research questions concerning what information a robot should collect, when sensing should occur, how long information should be retained, where it should be processed, and how information about users and bystanders should be protected. Relevant directions include data minimization, context-aware sensing, on-device processing, access control, and mechanisms that make the robot's sensing and data use understandable to users \cite{bell2025always}. These considerations are especially important for systems whose behavioral adaptation depends on accumulating experience over extended periods.

\section{Related work}

Our proposal sits at the intersection of two established research traditions: animal-inspired social robotics and modern robot learning.

Animal-like robotic systems have long explored how perception, internal state, and behavior-selection mechanisms can produce sustained interaction. Sony AIBO incorporated learning and developmental mechanisms that altered behavior over time \cite{fujita2001aibo}, while MiRo used a biomimetic control architecture to coordinate competing, context-sensitive behaviors in real time \cite{prescott2017miro}. LOVOT similarly employs a designed repertoire of continual and reflexive behaviors to support coherent social interaction \cite{yoshida2022production}. These systems demonstrate how perception and behavior selection mechanisms can produce animal-like interactions. However much of that behavior remains largely shaped by engineered behavioral architectures.

In parallel, recent robot-learning research has substantially advanced the acquisition and generalization of embodied skills. Vision-language-action models such as RT-2 \cite{zitkovich2023rt2} and OpenVLA \cite{kim2025openvla} learn mappings from multimodal observations and semantic context to robot actions. Systems such as QuarVLA extend related approaches to quadruped locomotion and manipulation \cite{ding2024quarvla}. World-model approaches such as DayDreamer \cite{wu2023daydreamer} learn adaptive sensorimotor behavior through interaction with the physical environment, and emerging world-action models jointly model future observations and actions \cite{ye2026worldactionmodels}.

The aforementioned two traditions emphasize different aspects of embodied competence. Traditional animal-like robotic systems have emphasized explicitly engineered behavioral repertoires and mechanisms for coordinating them, whereas modern robot learning has substantially
advanced the acquisition of individual skills, representations, predictive models, and generalization across tasks and environments. In this paper, we examine the following question: can modern robot learning methods acquire and coordinate the kind of persistent,
mid-level behavioral repertoire commonly observed in animals? See figure \ref{fig:ethological-layers}.

\section{Discussion}

Progress in robot learning is steadily expanding the range of tasks that embodied agents can perform. The perspective developed in this paper suggests that another form of competence deserves comparable attention: the ability to remain continuously and appropriately engaged with the physical and social environment. Behaviors such as attentional orienting, exploration, hazard avoidance, proximity regulation, engagement, and disengagement are individually simple compared with manipulation, planning, or instruction following. Their significance lies in their coordinated expression over time. An embodied agent must continually determine what deserves attention, which behavior should dominate, when an ongoing behavior should be interrupted, and how its actions should change as the surrounding context changes.

This motivates viewing the ethological behavioral substrate as a mid-level behavioral operating layer between low-level sensorimotor control and higher-level task competence. Such a layer need not determine how every behavior is implemented. Rather, it concerns how behavioral capabilities are organized and coordinated so that the agent maintains a coherent relationship with its surroundings while pursuing other objectives. Higher-level capabilities such as planning, manipulation, instruction following, and human--robot collaboration could then operate alongside persistent mechanisms for environmental awareness, behavioral regulation, and social responsiveness. An important research question is whether maintaining this competence alongside increasingly capable task policies improves robustness, adaptation, learning, and sustained interaction.

When considering systems in which relatively basic perceptual, regulatory, and social behaviors are continuously coordinated under changing conditions, animals emerge as a natural point of reference. However, artificial agents need not reproduce animal behavior directly; the behaviors they require may differ according to their morphology, environment, and role. The relevant question is therefore not whether a robot imitates an animal in detail, but whether it can develop the functional competence needed to remain situated, responsive, and behaviorally coherent over time.

This perspective also shifts how progress should be evaluated. Success cannot be characterized only by whether an agent can execute individual behaviors or complete externally specified tasks. Evaluation must also consider how behaviors interact under competing demands, whether responses remain appropriate as physical and social conditions change, and whether behavioral coherence persists across extended interaction. These questions are compatible with many computational approaches, including hierarchical policies, persistent memory and internal state, vision-language-action models, world models, and combinations of learned and engineered components. The central empirical question is whether organizing such mechanisms around persistent behavioral competence provides a useful foundation for increasingly capable embodied agents that can adapt to and coexist within dynamic physical and social environments.

\section*{GenAI Usage Disclosure}

The manuscript language was refined using generative AI assistance to improve clarity and readability. Generative AI was additionally used to create the figures included in the paper. The paper’s core ideas, system design decisions, technical content, and claims remain the responsibility of the authors.

\bibliographystyle{ACM-Reference-Format}
\bibliography{references.bib}

\end{document}